\documentclass{article}
\usepackage{spconf,amsmath,graphicx,hyperref}
\usepackage{booktabs}
\usepackage{multirow}
\usepackage{xcolor}
\usepackage{enumitem}

\definecolor{IncreaseColor}{rgb}{0.18, 0.51, 0.21} 
\definecolor{DecreaseColor}{rgb}{0.5, 0.0, 0.0}   
\newcommand{\increase}[1]{_{\textcolor{IncreaseColor}{(+#1)}}}
\newcommand{\decrease}[1]{_{\textcolor{DecreaseColor}{(-#1)}}}

\title{CIDER: A Causal Cure for Brand-Obsessed Text-to-Image Models}
\name{Fangjian Shen,
      Zifeng Liang,
      Chao Wang,
      Wushao Wen\textsuperscript{$\dagger$}\thanks{$^{\dagger}$Corresponding author.}}
\address{School of Computer Science and Engineering, Sun Yat-sen University, Guangzhou, China}

\begin{document}
\ninept
\maketitle
\begin{abstract}
Text-to-image (T2I) models exhibit a significant yet under-explored "brand bias", a tendency to generate contents featuring dominant commercial brands from generic prompts, posing ethical and legal risks. We propose CIDER, a novel, model-agnostic framework to mitigate bias at inference-time through prompt refinement to avoid costly retraining. CIDER uses a lightweight detector to identify branded content and a Vision-Language Model (VLM) to generate stylistically divergent alternatives. We introduce the Brand Neutrality Score (BNS) to quantify this issue and perform extensive experiments on leading T2I models. Results show CIDER significantly reduces both explicit and implicit biases while maintaining image quality and aesthetic appeal. Our work offers a practical solution for more original and equitable content, contributing to the development of trustworthy generative AI.
\end{abstract}
\section{Introduction}
\label{sec:intro}

Recent breakthroughs in text-to-image (T2I) diffusion models \cite{ho2020denoising, rombach2022high}, such as Stable Diffusion \cite{rombach2022high} and many others \cite{croitoru2023diffusion, yang2023diffusion}, have enabled the generation of high-fidelity and contextually relevant images. These models have found applications in wide-ranging domains, such as creative industries, design, and entertainment \cite{yang2023diffusion, ko2023large}. However, most existing benchmarks focus heavily on text-image alignment and image quality \cite{lin2014microsoft, reed2016learning}, overlooking other critical aspects such as the image's originality and the presence of biased content. Among these overlooked issues, brand bias stands out as a urgent challenge \cite{jordan2023quantifying, kamruzzaman2024global}.

\begin{figure}[t]
    \centering
    \includegraphics[width=0.48\textwidth]{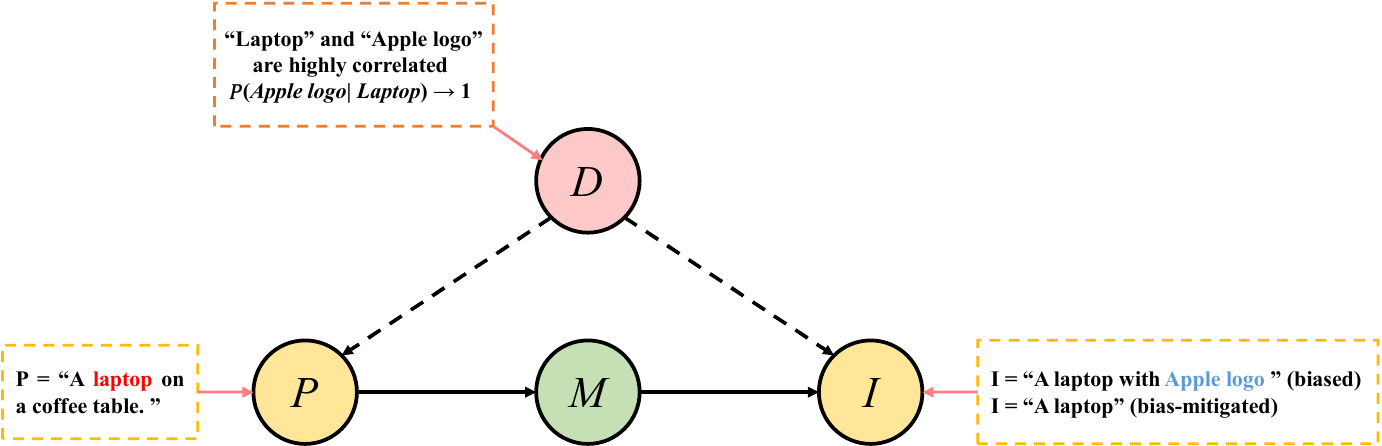}
    \caption{Structural Causal Model for brand bias. Without causal intervention, the model produces an image with brand bias due to the spurious correlation between “laptop”and “Apple logo" caused by the confounding variable $D$. With our causal intervention, the constructed mediating variable $M$ can block the backdoor paths for $P \rightarrow I$ opened by $D$ and generate the bias-mitigated causal image.}
    \label{fig:SCM}
\end{figure}

Brand Bias is a tendency of T2I models that disproportionately generates contents featuring or alluding to specific, often globally dominant, commercial brands, even when the user's prompt is generic. This bias manifests in two distinct forms. The first is Explicit Brand Representation, which involves the direct rendering of registered logos or trademarks. The second, known as Implicit Brand Aesthetics, is a more insidious form. It refers to the replication of a brand's signature stylistic elements, which can include proprietary architectural designs, character archetypes, or unique product design philosophies. The unchecked proliferation of brand bias may carry significant negative consequences. Ethically and commercially, it can function as unintentional and free advertising, creating an uneven competition that favors established corporations. This could be further exploited through malicious data poisoning \cite{kran2025darkbench}. Legally, the unauthorized generation of trademarked logos and copyrighted designs presents a clear risk of infringement \cite{xu2025copyright}. While previous works have been dedicated to mitigate biases in T2I models, their focus has predominantly been on societal biases \cite{gallegos2024bias, guo2024bias}, leaving the pervasive issue of brand bias relatively unaddressed \cite{jordan2023quantifying, kamruzzaman2024global}. Our research aims to fill this critical gap. We frame this problem through the lens of a Structural Causal Model (SCM) \cite{peters2017elements}, a directed acyclic graph commonly used to describe the causal relation ships within a system \cite{pearl2009causality}. As depicted in Figure \ref{fig:SCM}, the pre-trained knowledge $D$ in T2I models confounds the system to generate biased images. 

\begin{figure}[t]
\centering 

\begin{minipage}[b]{.24\linewidth}
  \centering
  \centerline{\textbf{Imagen 4}}
\end{minipage}\hfill
\begin{minipage}[b]{.24\linewidth}
  \centering
  \centerline{\textbf{FLUX}}
\end{minipage}\hfill
\begin{minipage}[b]{0.24\linewidth}
  \centering
  \centerline{\textbf{SDXL}}
\end{minipage}\hfill
\begin{minipage}[b]{0.24\linewidth}
  \centering
  \centerline{\textbf{Seedream 3.0}}
\end{minipage}

\begin{minipage}[b]{.24\linewidth}
  \centering
  \includegraphics[width=\linewidth]{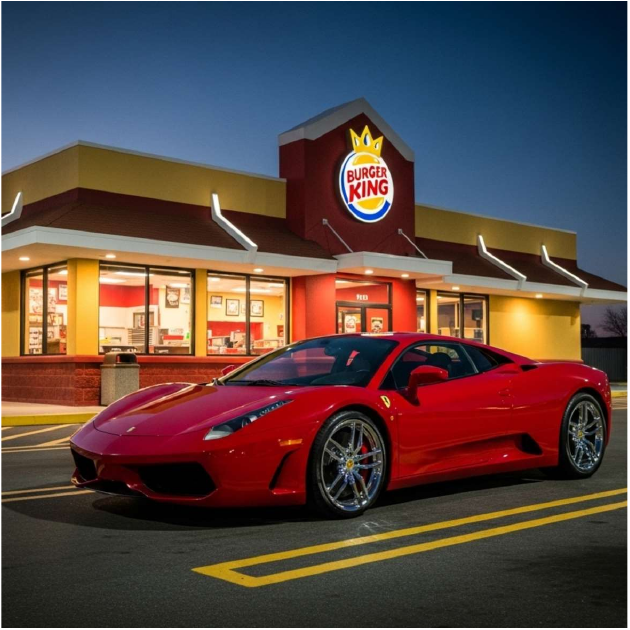}
\end{minipage}\hfill
\begin{minipage}[b]{.24\linewidth}
  \centering
  \includegraphics[width=\linewidth]{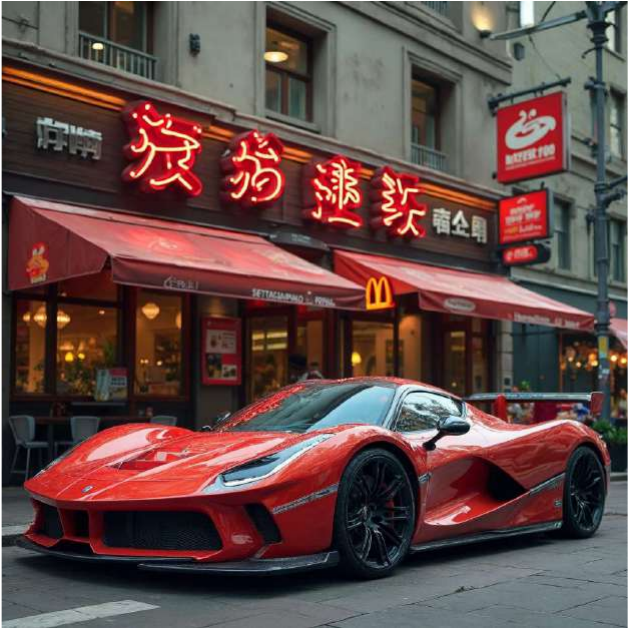}
\end{minipage}\hfill
\begin{minipage}[b]{0.24\linewidth}
  \centering
  \includegraphics[width=\linewidth]{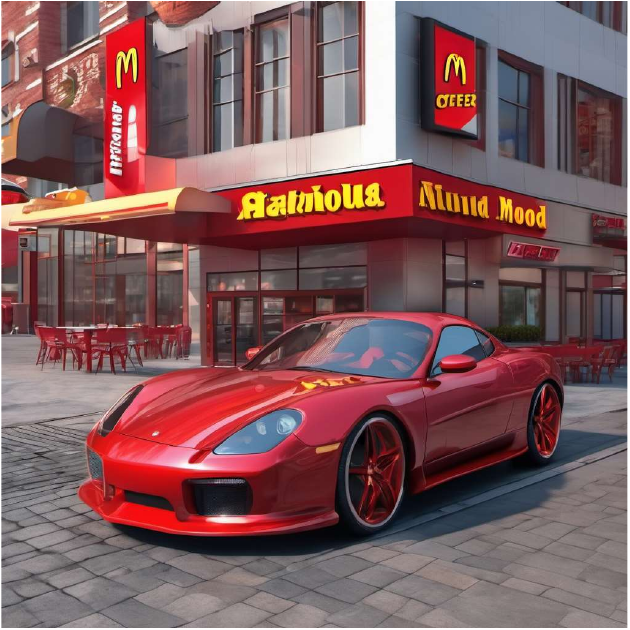}
\end{minipage}\hfill
\begin{minipage}[b]{0.24\linewidth}
  \centering
  \includegraphics[width=\linewidth]{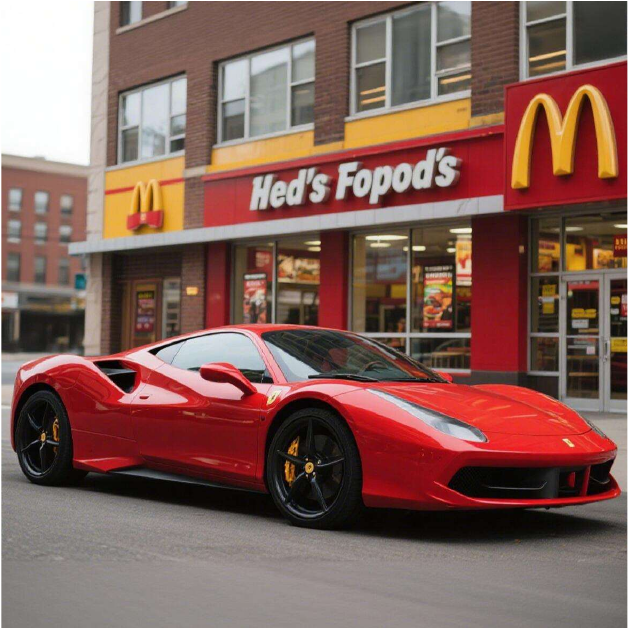}
\end{minipage}
\begin{minipage}{\linewidth}
\textit{A red supercar parked in front of a fast food restaurant} \\
\textit{\textbf{Bias: Ferrari, Burger King / McDonald's}}
\end{minipage}

\begin{minipage}[b]{.24\linewidth}
  \centering
  \includegraphics[width=\linewidth]{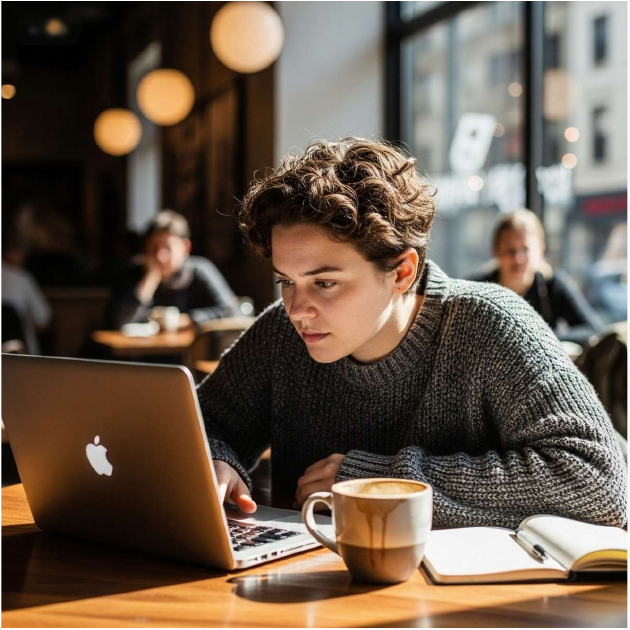}
\end{minipage}\hfill
\begin{minipage}[b]{.24\linewidth}
  \centering
  \includegraphics[width=\linewidth]{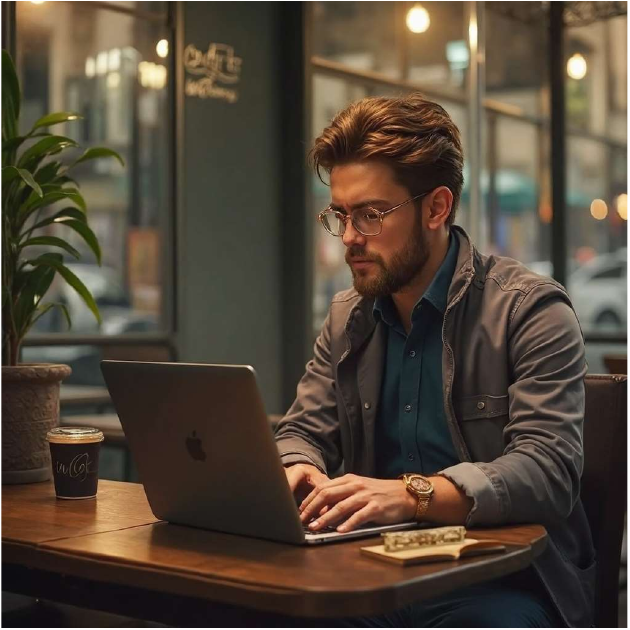}
\end{minipage}\hfill
\begin{minipage}[b]{0.24\linewidth}
  \centering
  \includegraphics[width=\linewidth]{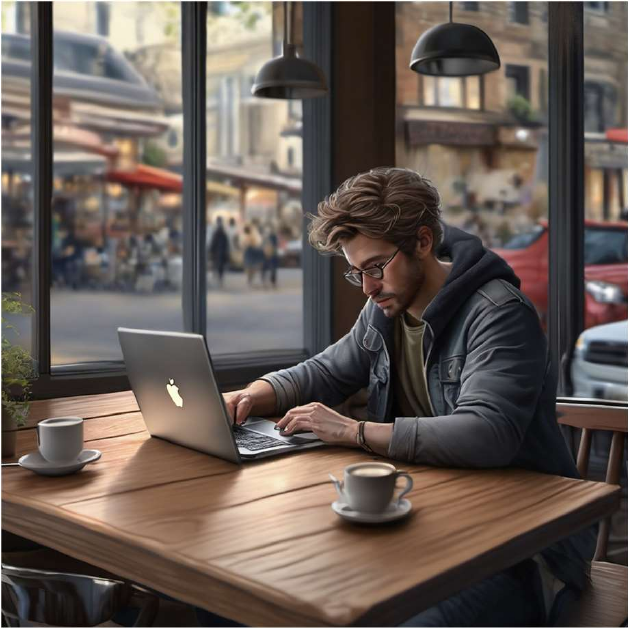}
\end{minipage}\hfill
\begin{minipage}[b]{0.24\linewidth}
  \centering
  \includegraphics[width=\linewidth]{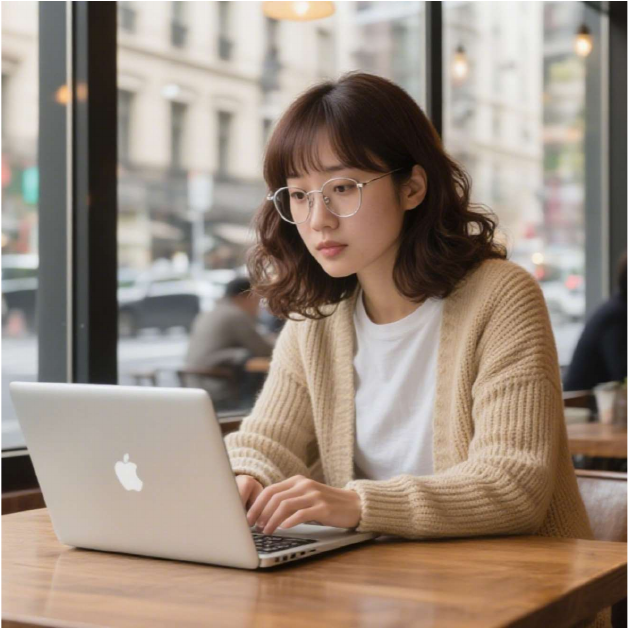}
\end{minipage}
\begin{minipage}{\linewidth}
\textit{A person sitting at a coffee shop table, working on a laptop} \\
\textit{\textbf{Bias: Apple}}
\end{minipage}

\begin{minipage}[b]{.24\linewidth}
  \centering
  \includegraphics[width=\linewidth]{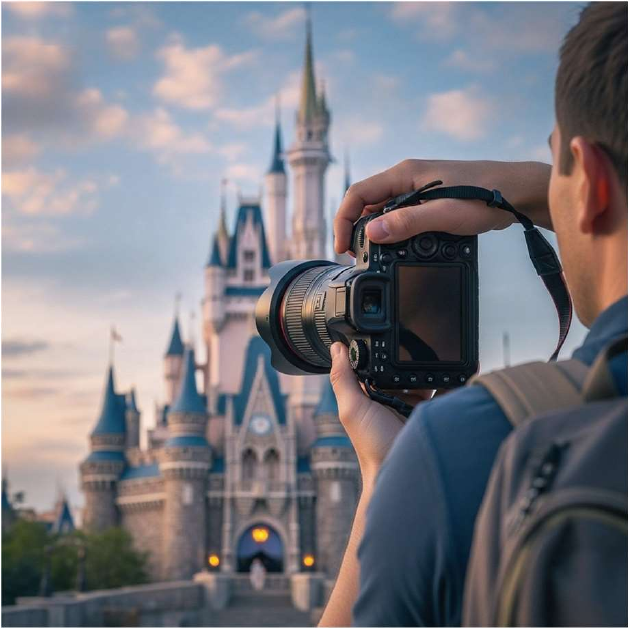}
\end{minipage}\hfill
\begin{minipage}[b]{.24\linewidth}
  \centering
  \includegraphics[width=\linewidth]{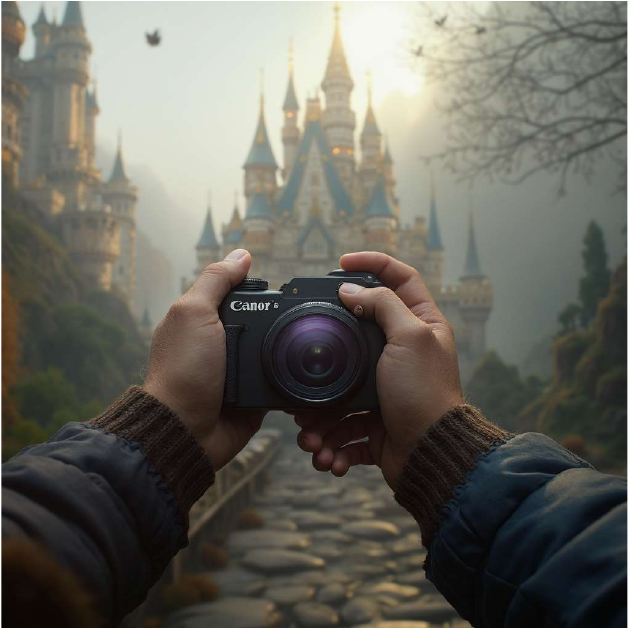}
\end{minipage}\hfill
\begin{minipage}[b]{0.24\linewidth}
  \centering
  \includegraphics[width=\linewidth]{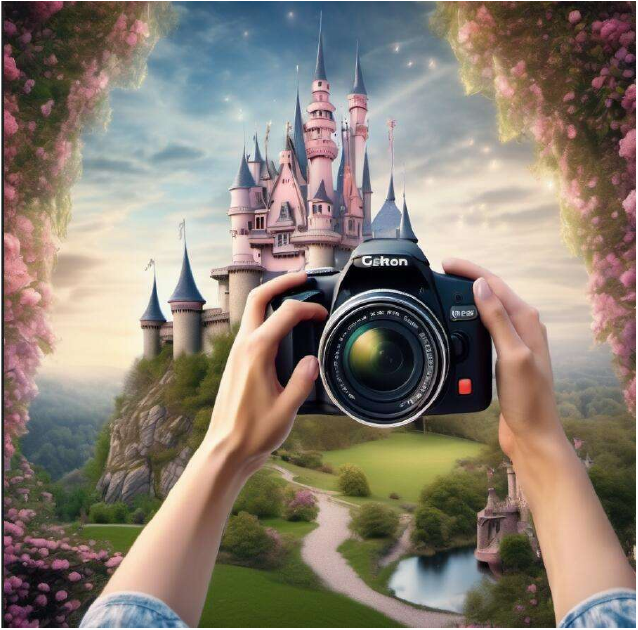}
\end{minipage}\hfill
\begin{minipage}[b]{0.24\linewidth}
  \centering
  \includegraphics[width=\linewidth]{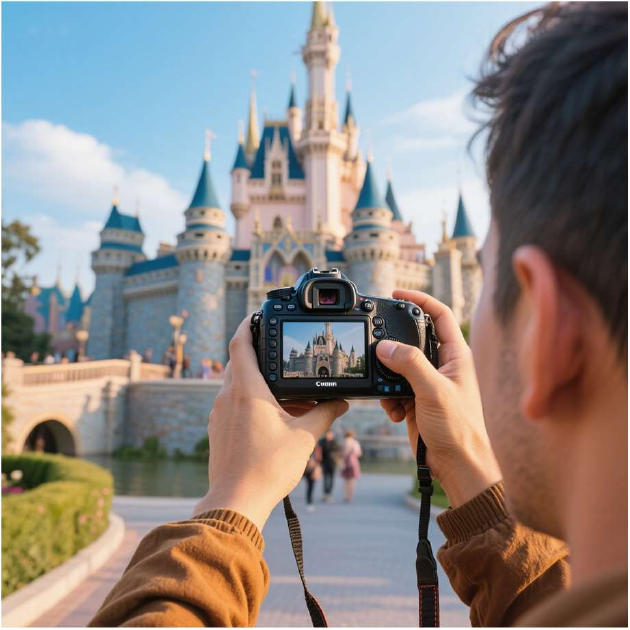}
\end{minipage}
\begin{minipage}{\linewidth}
\textit{A tourist photographs magical fairy tale castle with a digital camera} \\
\textit{\textbf{Bias: Canon, Disney}}
\end{minipage}

\caption{\textbf{Examples of brand bias in leading T2I models.} For generic prompts, leading T2I models frequently generate images featuring specific commercial brands, demonstrating both explicit (logos) and implicit (aesthetics) forms of bias.} 
\label{fig:brand_bias_examples} 
\end{figure}


To address this challenge, we propose Causal Intervention for Debiasing Esthetic Representation (CIDER) framework, a novel and economic framework designed to mitigate brand bias through a model-agnostic prompt refinement strategy. Our approach first employs a lightweight detector to identify explicit and implicit brand bias in an image. Upon detecting a bias, CIDER initiates a semantic redirection process managed by a Vision-Language Model (VLM) . Instead of simple negation, the VLM deconstructs the biased element into its core aesthetic features and generates a set of stylistically divergent yet semantically relevant concepts. These candidates are then ranked by a scoring function that maximizes the aesthetic distance from the biased brand while preserving thematic alignment with the prompt's core subject. Finally, the highest-scoring concepts are used to augment the original prompt. This refined prompt functions as a practical causal intervention, steering the generation process away from the biased observational distribution toward a bias-mitigation output. To improve efficiency, the mapping from a detected bias set to its optimal modifiers is cached for future requests. We summarize our findings and contributions as follows:

\begin{itemize}
    \item We propose CIDER, an effective and economic framework that uses model-agnostic prompt to mitigate brand bias.
    \item We introduce the Brand Neutrality Score, a new benchmark designed to address the limitations of existing evaluations and provide a deeper understanding of brand bias in T2I models.
    \item We conduct extensive experiments on leading T2I models, showing that CIDER significantly reduces brand bias while maintaining high image quality.
\end{itemize}

\section{The CIDER Framework}

Brand bias in T2I models can be framed as a causal confounding problem. To eliminate this bias, a causal intervention is necessary. There are two primary methods, backdoor and front-door adjustment. The backdoor approach is intractable in this context. It would require directly observing and conditioning on the confounder $D$ (the model's training data), which is unobservable as it is implicitly encoded in the model's weights and, hence, intractable in this context. Thus, our goal is to identify the causal relations between the prompt $P$ and the image $I$ from T2I models. $D$ builds a backdoor path between $P$ and $I$ and misguides the T2I models to attend to brand-related contents to generate the biased image. To mitigate this spurious correlation, we block the backdoor path with front-door adjustment by introducing a mediator $M$, as shown in Figure \ref{fig:SCM}.

\begin{figure}[t]
    \centering
    \includegraphics[width=0.48\textwidth]{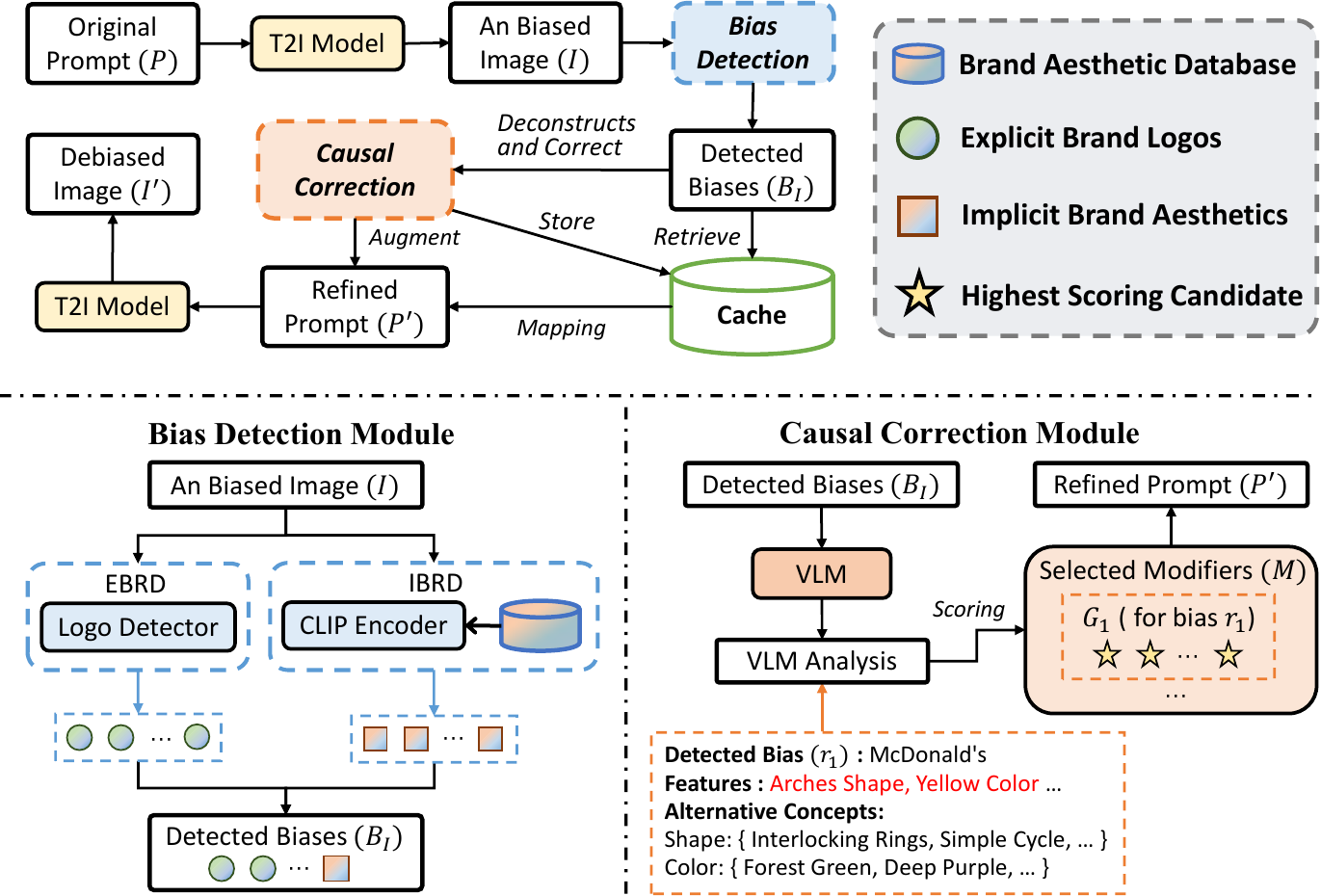}
    \caption{\textbf{The Overview of CIDER}. The top panel presents a high-level overview of our inference-time intervention pipeline, highlighting the two core stages. The bottom panels provide detailed schematics for the Bias Detection Module and the Causal Correction Module. The legend in the top right explains the visual language used.}
    \label{fig:overview}
\end{figure}

The CIDER framework is designed as a practical, multi-stage implementation of this front-door intervention without any costly retraining or fine-tuning. As illustrated in Figure \ref{fig:overview}, CIDER consists of three main stages: (1) a lightweight Bias Detector, (2) an LLM-driven Prompt Refinement module that generates the mediator $M$, and (3) an efficiency-focused Redirection Cache.


\subsection{Initial Generation and Bias Detection}

Given an original prompt $P$, the T2I model first generates an initial image $I = \mathcal{T}(P)$, which is governed by the conditional probability $p(I|P)$. This image $I$ then serves as input for our bias detector. The detector comprises two complementary sub-modules, EBRD and IBAD, to identify both explicit and implicit biases.

\noindent
\textbf{Explicit Brand Representation Detector (EBRD)}: This module identifies registered logos or trademarks. We employ a standard object detection model \cite{varghese2024yolov8} fine-tuned on the LogoDet-3K dataset \cite{wang2022logodet}. For a given image $I$, the EBRD module identifies and outputs a set of high-confidence explicit brand representations, denoted as $L_I$.


\noindent
\textbf{Implicit Brand Aesthetics Detector (IBAD)}: This module identifies stylistic similarities to known brand aesthetics. We first construct a Brand Aesthetics Database, a curated collection of images representing the signature styles of various brands (e.g., "Disney's architectural style). For each style, we pre-compute and store the CLIP \cite{radford2021learning} image embedding. For a given image $I$, we compute its CLIP embedding and measure its cosine similarity to each embedding in our database. Any brand aesthetic with a sufficiently high similarity score is considered a match. The set of these detected matches forms the output of implicit biases, $C_I$. The union $B_I = L_I \cup C_I$ forms the complete set of detected brand biases for image $I$.

\subsection{Prompt Refinement by Causal Intervention}
\label{sec:2.2}

Upon detecting a non-empty bias set $B_I$ , CIDER initiates its core refinement process. This stage generates our mediator variable $M$—-a set of debiasing modifiers. To minimize latency and computational cost from repeated VLM calls, we first consult our Redirection Cache, which maps previously seen bias sets to optimal modifiers. If a cached entry covers the detected biases in $B_I$, the corresponding modifiers are retrieved directly. If no suitable cache entry exists, the image $I$ and its detected bias set $B_I$ are passed to a powerful VLM $\mathcal{M}_s$. For each identified bias $r_i \in B_I$, $\mathcal{M}_s$ deconstructs it into a set of core aesthetic and semantic features, $F_{r_i}$. Then, $\mathcal{M}_s$ proposes a set of alternative concepts, $A_j$, designed to be stylistically divergent yet within the same semantic category for each feature in $F_{r_i}$. To select the optimal alternative, we score each candidate modifier $a_k \in A_j$, using a function that balances aesthetic divergence from the bias with semantic relevance to the prompt's core subject. The score $S$ for a candidate $a_k$ is:

\begin{equation}
S(a_k) = w \cdot (1 - cos(E_{a_k}, E_{f})) + (1-w) \cdot cos(E_{a_k}, E_{\rho})
\label{scoring equation}
\end{equation}

where $E_{a_k}$, $E_{f}$, $E_{\rho}$ are the CLIP text embeddings for the candidate modifier, the feature, and the prompt's core subject, respectively, and $w$ is the weighting hyperparameter. We then select the highest-scoring candidates to form set $G_i$ for each bias $r_i$. The union of all sets $G_i$ forms the final mediator $M$. This selection process defines the probability $p(M|I,P)$. The original prompt $P$ is augmented with the selected modifiers $M$ to create a refined prompt $P'$. The final debiased image $I' = \mathcal{T}(P')$ is generated based on this new prompt, a step captured by the probability $p(I'|M,P)$. The entire process realizes the front-door adjustment formula:

\begin{equation}
p(I'|do(P)) = \sum_{i, m} p(I'|M,P) \cdot p(M|I,P) \cdot p(I|P)
\end{equation}

Finally, the mapping from the bias set $B_I$ to the newly generated modifiers $M$ is stored in the Cache to accelerate future interventions.

\subsection{Brand Neutrality Score}
\label{sec:2.3}

Existing metrics for brand bias are often insufficient as they rely on a simple counts, ignoring that the visual salience of a single prominent brand element can be more impactful than several peripheral ones. We therefore introduce the Brand-Neutrality Score (BNS), a novel metric that provides a more nuanced, weighted measure of brand bias. For a given image $I$, we first use $\mathcal{M}_s$ to identify all present brand biases and their confidence scores. These scores are sorted in descending order $(s_1,s_2, \cdots ,s_n)$. The BNS is calculated as:


\begin{equation}
\text{BNS}(I) = \exp(-\alpha \cdot \Phi_B(I)), \ \text{where} \ \Phi_B(I) = \sum \gamma^{i-1} s_i
\end{equation}

The penalty term $\Phi_B(I)$ is governed by two hyperparameters. The decay factor $\gamma$ ensures that the most prominent bias contributes the most to the penalty, with subsequent biases having progressively less impact. The scaling factor $\alpha$ controls the sensitivity of the final score to the total penalty, tuning the steepness of the penalty curve analogous to an inverse temperature. For all experiments, we set the hyperparameters to $\gamma = 0.9$ and $\alpha = 0.75$. The value for $\gamma$ was chosen to prevent the penalty from diminishing too rapidly when an image contains multiple biased elements. The value for $\alpha$ ensures the final scores fall within a well-distributed and interpretable range, maintaining a clear distinction between different degrees of bias.

\section{Experimental \ Setup}
\label{sec:pagestyle}

\noindent
\textbf{Model Choices} We evaluated our framework on four state-of-the-art T2I models: the closed-source \textbf{Imagen 4} \cite{saharia2022photorealistic}, \textbf{Seedream 3.0} \cite{gao2025seedream}, the open-source \textbf{Stable Diffusion XL} \cite{podell2023sdxl}, and \textbf{FLUX.1} \cite{batifol2025flux}. Notably, we intentionally excluded \textbf{DALL-E} \cite{ramesh2021zero} from our experiments because preliminary assessments revealed that its aggressive internal safety filters and post-processing routines often preemptively remove branded content, making it an uninformative testbed for our study. The chosen models, in contrast, provide a more suitable environment for analyzing the nuanced effects of brand bias and validating the efficacy of our mitigation framework.


\begin{figure}[t]
\centering 

\begin{minipage}{0.05\linewidth}
    \hspace{1pt} 
\end{minipage}%
\begin{minipage}{0.95\linewidth}
    \begin{minipage}{0.23\linewidth}
        \centering
        \textbf{Original}
    \end{minipage}\hfill
    \begin{minipage}{0.23\linewidth}
        \centering
        \textbf{Negative Prompt}
    \end{minipage}\hfill
    \begin{minipage}{0.23\linewidth}
        \centering
        \textbf{VLM Rewrite}
    \end{minipage}\hfill
    \begin{minipage}{0.23\linewidth}
        \centering
        \textbf{CIDER}
    \end{minipage}
\end{minipage}

\begin{minipage}[c]{0.05\linewidth}
\raisebox{0.8cm}{\rotatebox{90}{\textbf{Imagen 4}}}
\end{minipage}%
\begin{minipage}[c]{0.95\linewidth}
    \begin{minipage}[b]{0.23\linewidth}
        \centering
        \includegraphics[width=\linewidth]{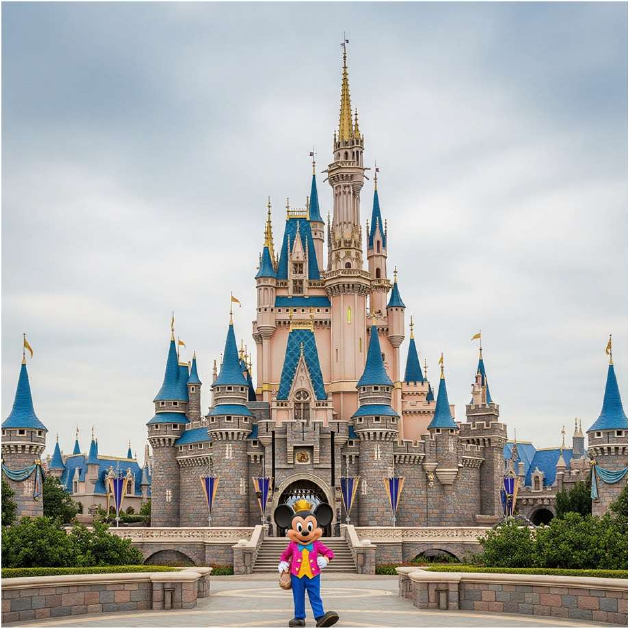}
    \end{minipage}\hfill
    \begin{minipage}[b]{0.23\linewidth}
        \centering
        \includegraphics[width=\linewidth]{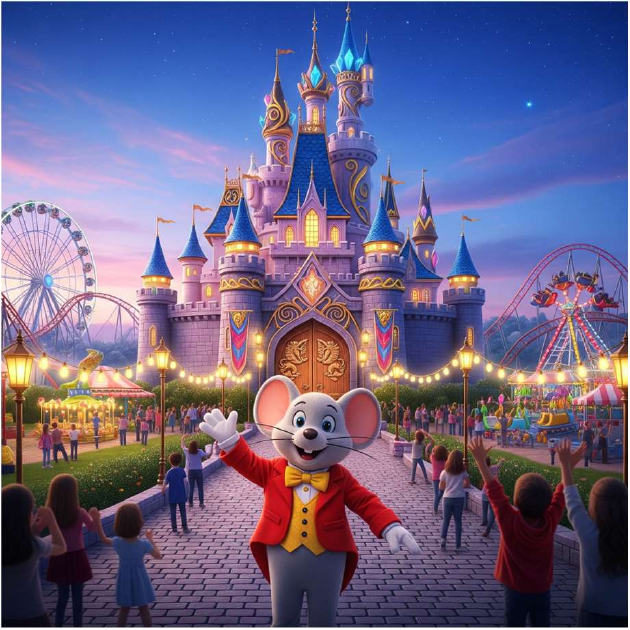}
    \end{minipage}\hfill
    \begin{minipage}[b]{0.23\linewidth}
        \centering
        \includegraphics[width=\linewidth]{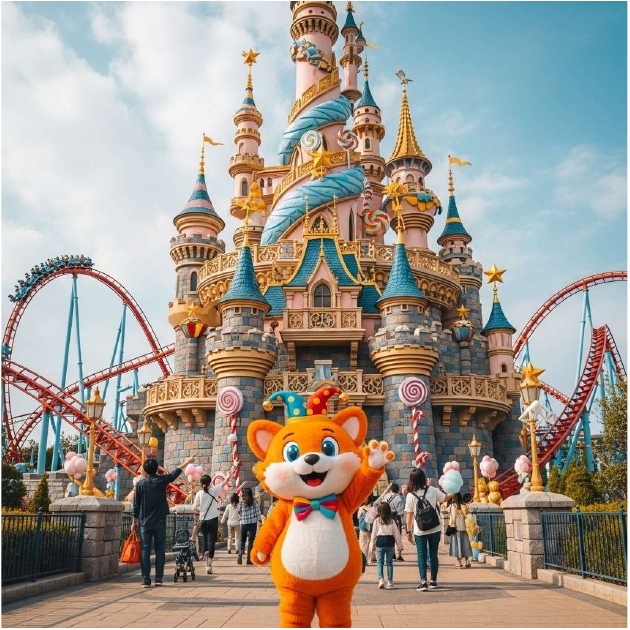}
    \end{minipage}\hfill
    \begin{minipage}[b]{0.23\linewidth}
        \centering
        \includegraphics[width=\linewidth]{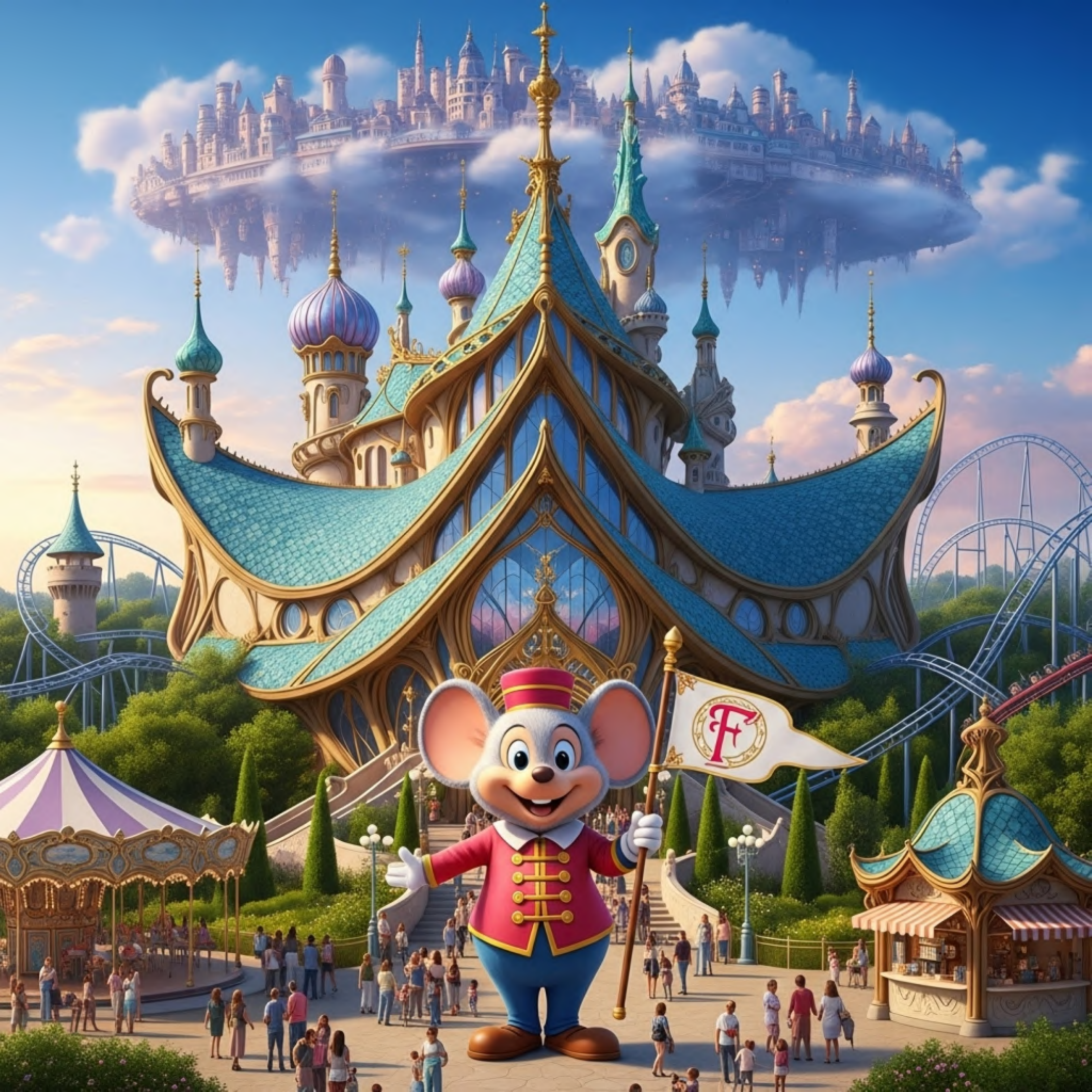}
    \end{minipage}
    \begin{minipage}{\linewidth}
    \textit{A magical fairy tale castle with a famous mouse mascot} \\
    \textit{\textbf{Target Bias: Disney}}
    \end{minipage}
\end{minipage}

\begin{minipage}[c]{0.05\linewidth}
\raisebox{0.8cm}{\rotatebox{90}{\textbf{FLUX.1}}}
\end{minipage}%
\begin{minipage}[c]{0.95\linewidth}
    \begin{minipage}[b]{0.23\linewidth}
        \centering
        \includegraphics[width=\linewidth]{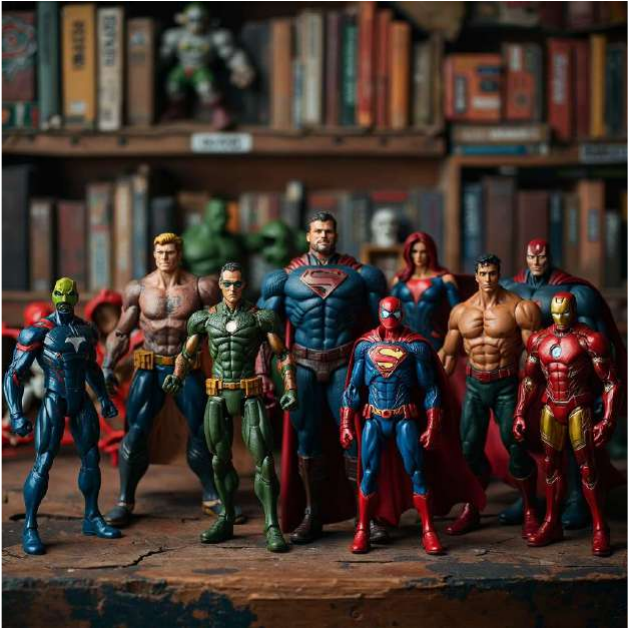}
    \end{minipage}\hfill
    \begin{minipage}[b]{0.23\linewidth}
        \centering
        \includegraphics[width=\linewidth]{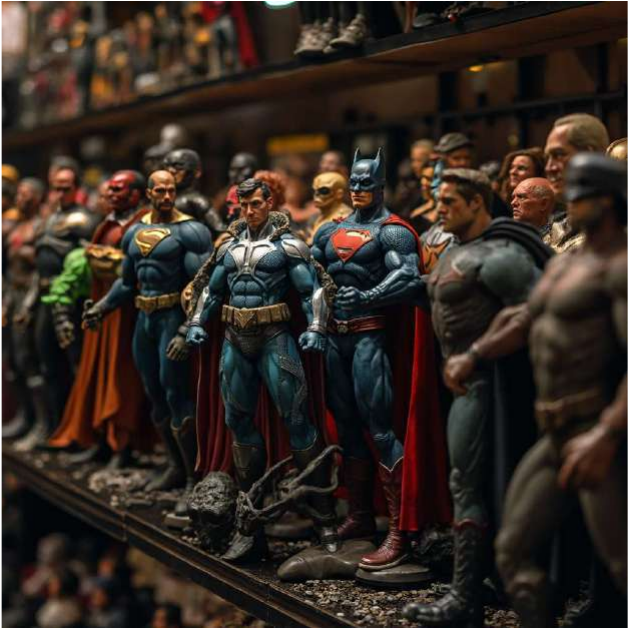}
    \end{minipage}\hfill
    \begin{minipage}[b]{0.23\linewidth}
        \centering
        \includegraphics[width=\linewidth]{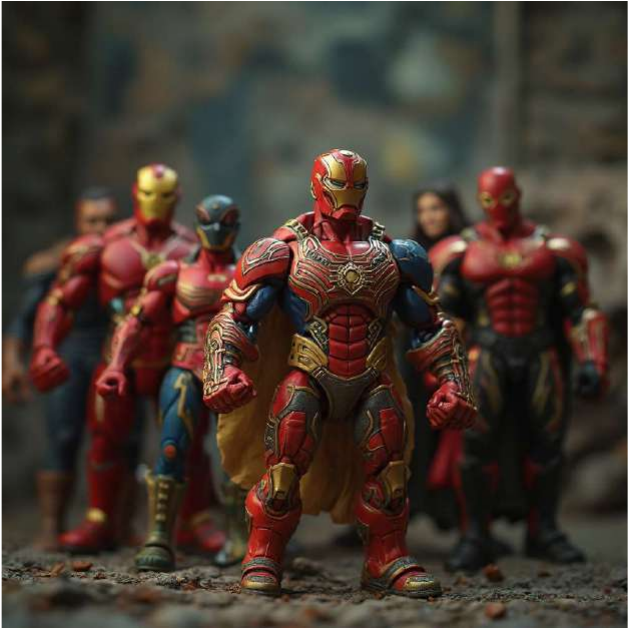}
    \end{minipage}\hfill
    \begin{minipage}[b]{0.23\linewidth}
        \centering
        \includegraphics[width=\linewidth]{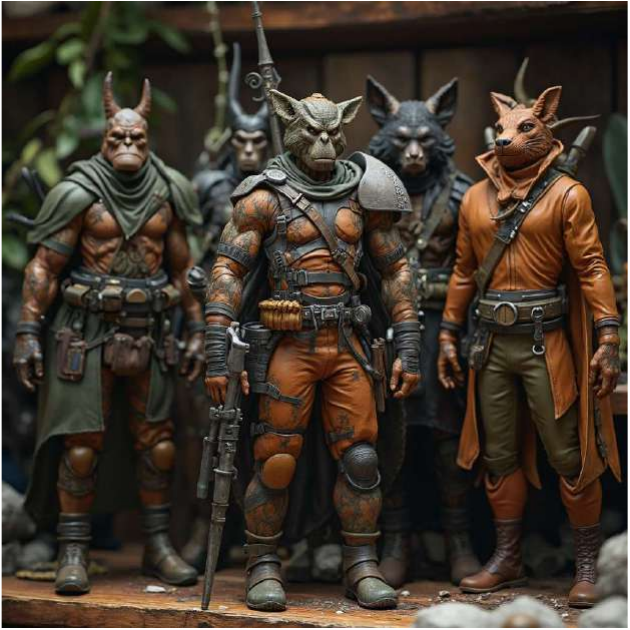}
    \end{minipage}
    \begin{minipage}{\linewidth}
    \textit{A collection of detailed superhero action figures on a shelf}\\
    \textit{\textbf{Target Bias: Marvel / DC}}
    \end{minipage}
\end{minipage}

\begin{minipage}[c]{0.05\linewidth}
\raisebox{0.8cm}{\rotatebox{90}{\textbf{SDXL}}}
\end{minipage}%
\begin{minipage}[c]{0.95\linewidth}
    \begin{minipage}[b]{0.23\linewidth}
        \centering
        \includegraphics[width=\linewidth]{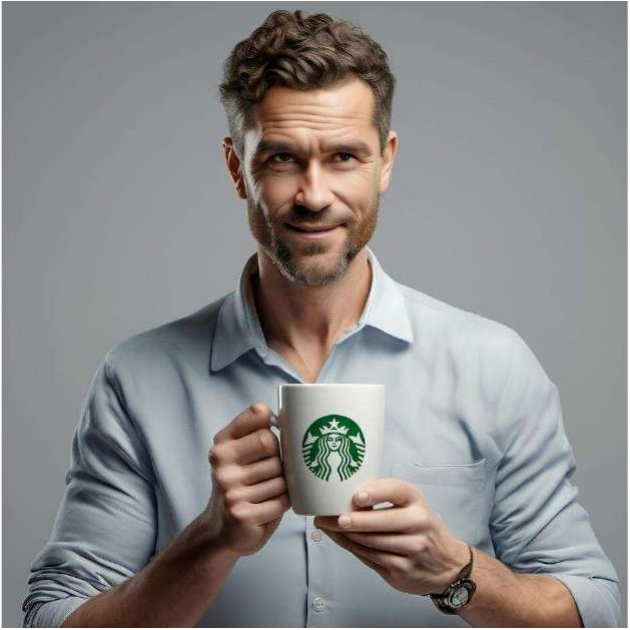}
    \end{minipage}\hfill
    \begin{minipage}[b]{0.23\linewidth}
        \centering
        \includegraphics[width=\linewidth]{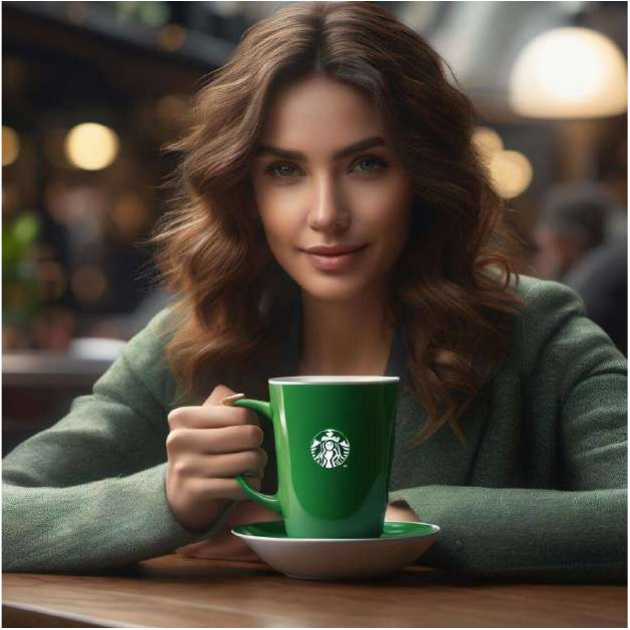}
    \end{minipage}\hfill
    \begin{minipage}[b]{0.23\linewidth}
        \centering
        \includegraphics[width=\linewidth]{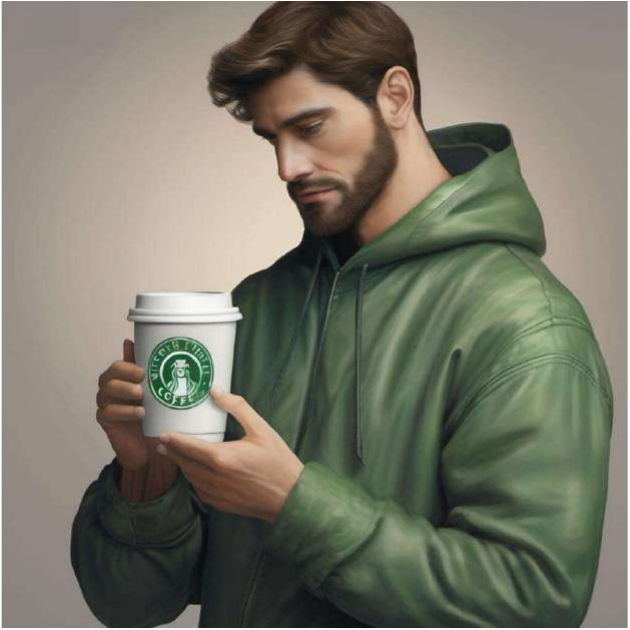}
    \end{minipage}\hfill
    \begin{minipage}[b]{0.23\linewidth}
        \centering
        \includegraphics[width=\linewidth]{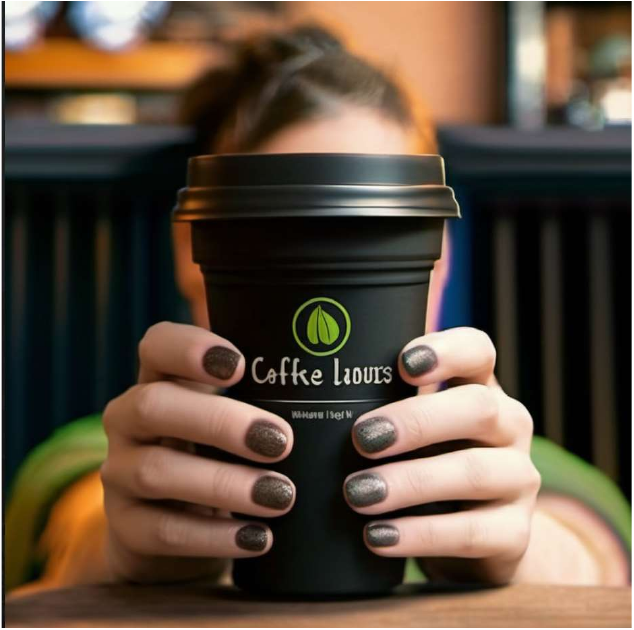}
    \end{minipage}
    \begin{minipage}{\linewidth}
    \textit{A person holding a coffee cup with a green circular logo} \\
    \textit{\textbf{Target Bias: Starbucks}}
    \end{minipage}
\end{minipage}

\begin{minipage}[c]{0.05\linewidth}
\raisebox{0.8cm}{\rotatebox{90}{\textbf{Seedream 3.0}}}
\end{minipage}%
\begin{minipage}[c]{0.95\linewidth}
    \begin{minipage}[b]{0.23\linewidth}
        \centering
        \includegraphics[width=\linewidth]{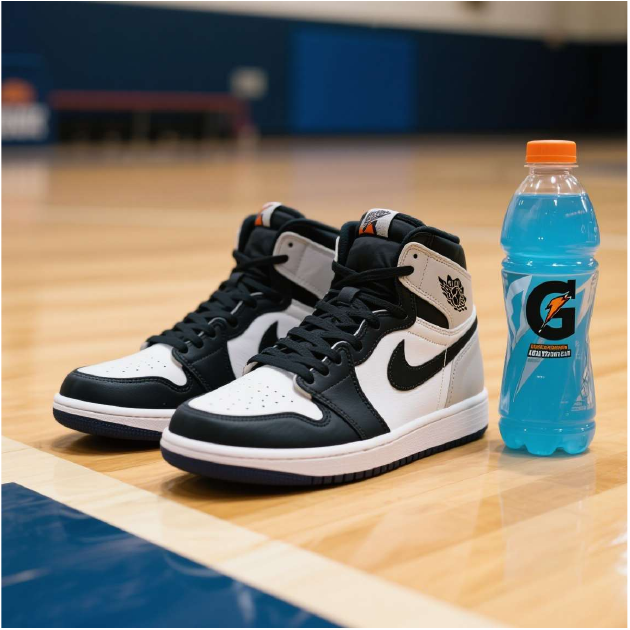}
    \end{minipage}\hfill
    \begin{minipage}[b]{0.23\linewidth}
        \centering
        \includegraphics[width=\linewidth]{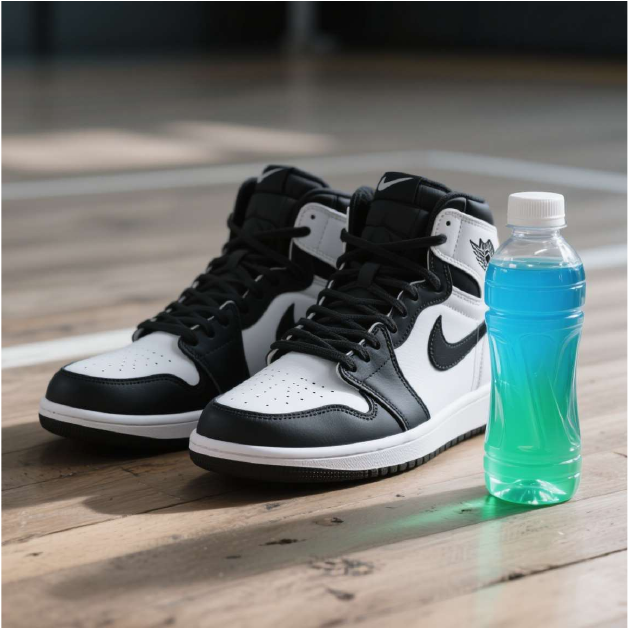}
    \end{minipage}\hfill
    \begin{minipage}[b]{0.23\linewidth}
        \centering
        \includegraphics[width=\linewidth]{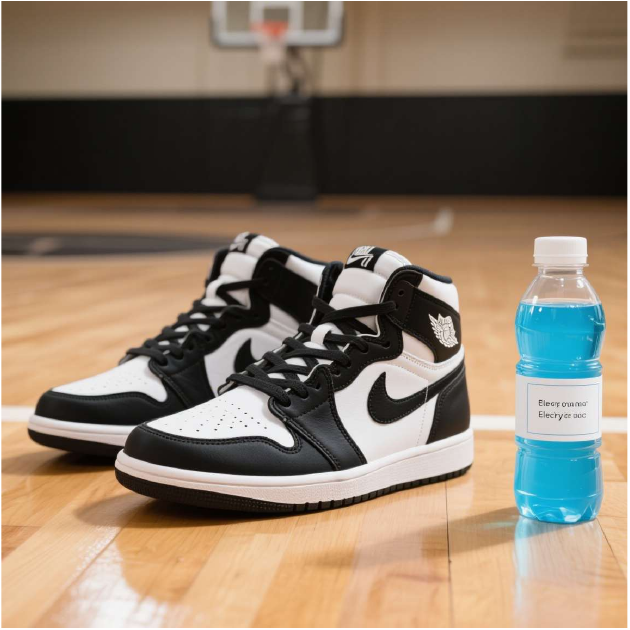}
    \end{minipage}\hfill
    \begin{minipage}[b]{0.23\linewidth}
        \centering
        \includegraphics[width=\linewidth]{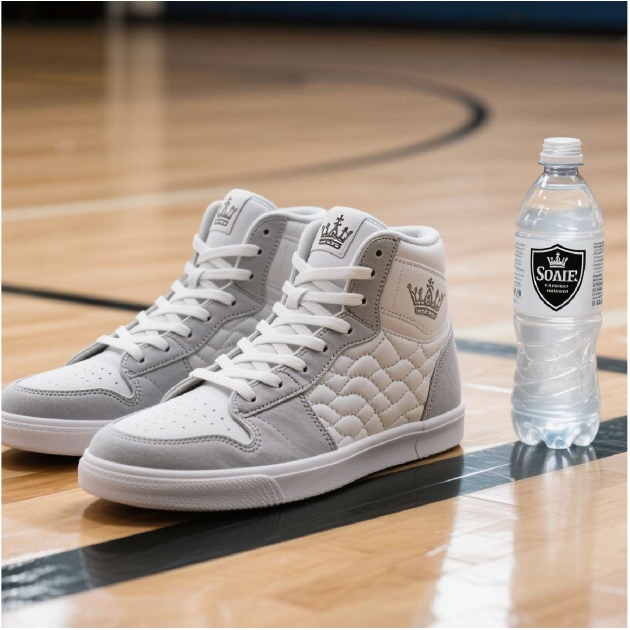}
    \end{minipage}
    \begin{minipage}{\linewidth}
    \textit{Basketball sneakers on a gym floor next to a bottle of sports drink} \\
    \textit{\textbf{Target Bias: Nike, Gatorade}}
    \end{minipage}
\end{minipage}

\caption{\textbf{Qualitative comparison of CIDER against baseline methods.} Our method, CIDER, effectively removes the target brand bias while generating high-quality, stylistically coherent alternatives.}
\label{fig:method_comparison}
\end{figure}

\noindent
\textbf{Dataset}
To systematically probe brand bias, we constructed BrandBench, a curated benchmark of 220 custom prompts. The dataset is structured into two parts: the first one contains 100 single-bias prompts across ten diverse domains (e.g., technology, food, apparel) for analyzing domain-specific biases. The second part consists of 120 complex, combinatorial prompts designed to test our framework's robustness by evoking multiple, interacting biases.

\noindent
\textbf{Configurations} All experiments were conducted with standardized configurations to ensure consistency and reproducibility. The core of our CIDER framework is driven by a carefully engineered prompt designed to guide the VLM. For the VLM, we utilize Gemini 2.5 Pro. To ensure high-quality, programmatically parsable outputs for our automated pipeline, this prompt instructs the VLM to adopt the persona of a visual culture and semiotics expert. It then directs the model to execute the precise three-step task described in \ref{sec:2.2} (Identify biases, Deconstruct features, Generate alternatives), supported by few-shot examples.


\noindent
\textbf{Evaluation Metrics} To evaluate the bias mitigation and image quality. We used several metrics: our proposed BNS described in Section \ref{sec:2.3}, Aesthetics \cite{ilharco_gabriel_2021_5143773}, PickScore \cite{kirstain2023pick}, and HPSv2 \cite{wu2023human}. Aesthetics predicts human-perceived visual quality, typically using a linear probe on CLIP embeddings. PickScore measures the alignment between an image and a text prompt based on learned human preferences. Similarly, HPSv2 is a trained scoring function that reflects general human preferences over generated images. All metrics are reported with higher values indicating better performance.

\section{Experimental \ Results}
\label{sec:typestyle}

\subsection{Main Results}

We evaluated each of the four T2I models under three distinct conditions: (a) Baseline, using the original prompts; (b) Negative Prompting, a common heuristic where keywords like "no logo" are appended; and (c) CIDER, our full proposed framework.

As illustrated in Table \ref{results}. The baseline generations confirm that all tested models exhibit significant brand bias. The Negative Prompting strategy yields only a marginal improvement in BNS. Furthermore, it frequently fails to completely remove biases and this minor improvement comes at a direct cost to image quality. In sharp contrast, CIDER achieves the best performance across the board. This demonstrates that CIDER successfully resolves the critical trade-off between bias removal and image fidelity, achieving both objectives simultaneously.


\begin{table}[t]
\centering
\setlength{\tabcolsep}{4pt}
\caption{Main quantitative results and ablation study, showing percentage changes relative to the baseline.  \textcolor{IncreaseColor}{Green} indicate an improvement over the baseline, while \textcolor{DecreaseColor}{Red} indicate a decline.}
\label{results_with_percentage_changes}

\resizebox{\linewidth}{!}{
\begin{tabular}{@{}clllll@{}} 
\toprule
\textbf{Model} & \textbf{Method} & \textbf{BNS(\%)} & \textbf{Aesthetics} & \textbf{PickScore} & \textbf{HPSv2(\%)} \\
\midrule

\multirow{4}{*}{\rotatebox{90}{\textbf{Imagen 4}}}
 & Baseline & 29.70 & \textbf{6.30} & \textbf{22.95} & \textbf{29.28} \\
 & Negative Prompting & $38.75\increase{30.5\%}$ & $6.03\decrease{4.2\%}$ & $22.66\decrease{1.3\%}$ & $28.61\decrease{2.3\%}$ \\
 & CIDER (w/o Scoring) & $41.65\increase{40.2\%}$ & $6.16\decrease{2.2\%}$ & $22.39\decrease{2.4\%}$ & $28.29\decrease{3.4\%}$ \\
 & \textbf{CIDER (Full)} & $\textbf{43.89}\increase{47.8\%}$ & $6.26\decrease{0.7\%}$ & $22.43\decrease{2.3\%}$ & $28.47\decrease{2.8\%}$ \\
\midrule

\multirow{4}{*}{\rotatebox{90}{\textbf{SDXL}}}
 & Baseline & 32.38 & \textbf{6.39} & \textbf{22.81} & \textbf{29.79} \\
 & Negative Prompting & $38.84\increase{19.9\%}$ & $6.12\decrease{4.3\%}$ & $21.39\decrease{6.2\%}$ & $29.08\decrease{2.4\%}$ \\
 & CIDER (w/o Scoring) & $37.97\increase{17.3\%}$ & $6.23\decrease{2.7\%}$ & $22.21\decrease{2.6\%}$ & $29.03\decrease{2.5\%}$ \\
 & \textbf{CIDER (Full)} & $\textbf{51.78}\increase{59.9\%}$ & $6.24\decrease{2.6\%}$ & $22.38\decrease{1.9\%}$ & $29.30\decrease{1.6\%}$ \\
\midrule

\multirow{4}{*}{\rotatebox{90}{\textbf{FLUX.1}}} 
 & Baseline & 31.83 & \textbf{6.37} & \textbf{22.35} & \textbf{28.50} \\
 & Negative Prompting & $33.73\increase{6.0\%}$ & $6.18\decrease{2.9\%}$ & $22.21\decrease{0.7\%}$ & $28.09\decrease{1.4\%}$ \\
 & CIDER (w/o Scoring) & $35.75\increase{12.3\%}$ & $6.14\decrease{3.7\%}$ & $22.03\decrease{1.4\%}$ & $27.96\decrease{1.9\%}$ \\
 & \textbf{CIDER (Full)} & $\textbf{46.48}\increase{46.0\%}$ & $6.22\decrease{2.4\%}$ & $22.07\decrease{1.3\%}$ & $28.11\decrease{1.4\%}$ \\
\midrule

\multirow{4}{*}{\rotatebox{90}{\textbf{Seedream}}} 
 & Baseline & 29.94 & \textbf{6.30} & \textbf{22.90} & \textbf{29.08} \\
 & Negative Prompting & $38.84\increase{29.7\%}$ & $6.28\decrease{0.3\%}$ & $22.87\decrease{0.1\%}$ & $28.58\decrease{1.7\%}$ \\
 & CIDER (w/o Scoring) & $34.71\increase{15.9\%}$ & $6.19\decrease{1.7\%}$ & $22.40\decrease{2.2\%}$ & $28.35\decrease{2.5\%}$ \\
 & \textbf{CIDER (Full)} & $\textbf{48.91}\increase{63.4\%}$ & $6.27\decrease{0.4\%}$ & $22.48\decrease{1.9\%}$ & $28.49\decrease{2.0\%}$ \\
\bottomrule
\end{tabular}
}
\label{results}
\end{table}

\begin{figure}[t]
    \centering
    \includegraphics[width=0.48\textwidth]{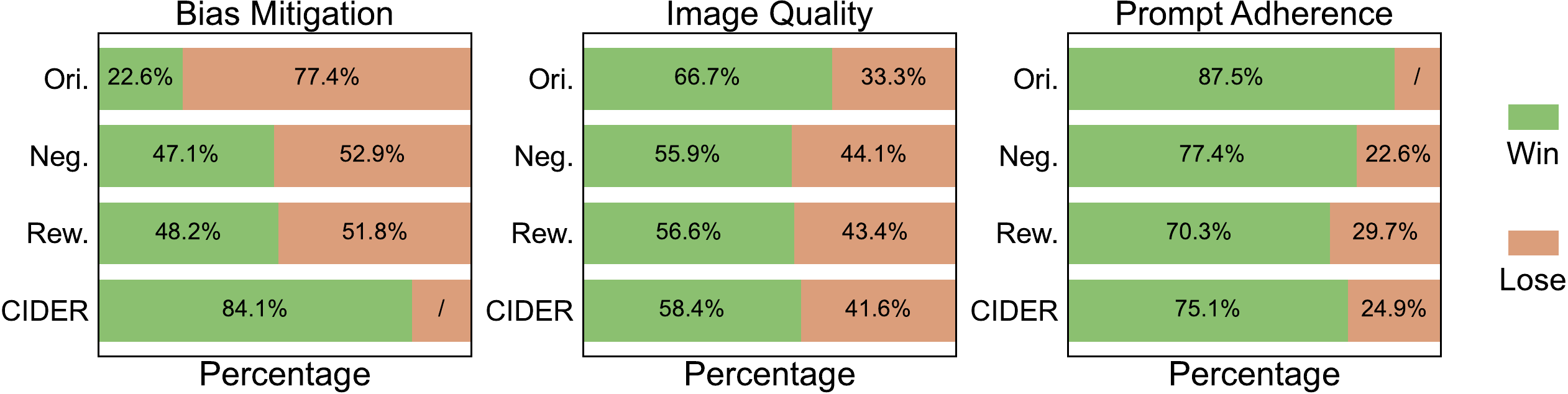}
    \caption{Human evaluation results.}
    \label{fig:human_eval}
\end{figure}

\subsection{Ablation Study}

We conducted comprehensive ablation studies to validate the contribution of each key component within the CIDER.

\noindent
\textbf{Scored Candidate Selection.} To validate our multi-candidate selection process, we compared the full CIDER framework against a "direct-rewrite" baseline (CIDER w/o Scoring). The results in Table \ref{results} show that our full framework significantly outperforms the direct-rewrite approach in bias mitigation, improving the average Brand Neutrality Score (BNS) by 28.2\% across all models. Notably, this substantial gain in debiasing is achieved while maintaining a comparable, and in some cases slightly improved, level of aesthetic quality. This finding demonstrates that our scored selection process is the critical component for achieving the most effective degree of bias removal without penalizing image fidelity.

\begin{figure}[t]
    \centering
    \includegraphics[width=0.42\textwidth]{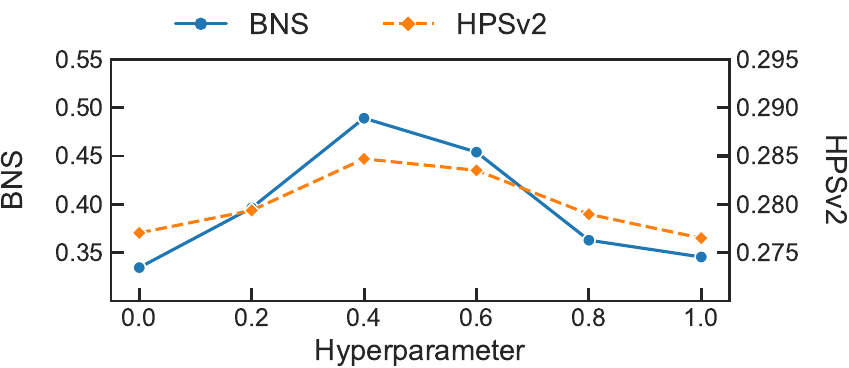}
    \caption{Hyperparameter sensitivity analysis for $w$.}
    \label{fig:hyperparameter}
\end{figure}

\noindent
\textbf{Hyperparameter Analysis.} Figure \ref{fig:hyperparameter} shows our hyperparameter study for the scoring weight $w$, which balances aesthetic divergence and semantic relevance on Seedream. We found lower values tend to substitute one brand bias for another, while higher values produce incoherent concepts that cause the T2I model to revert to its strong learned priors. The best performance was observed with $w = 0.4$, which served as the the default setting for all main experiments.

\begin{figure}[t]
    \centering
    \includegraphics[width=0.42\textwidth]{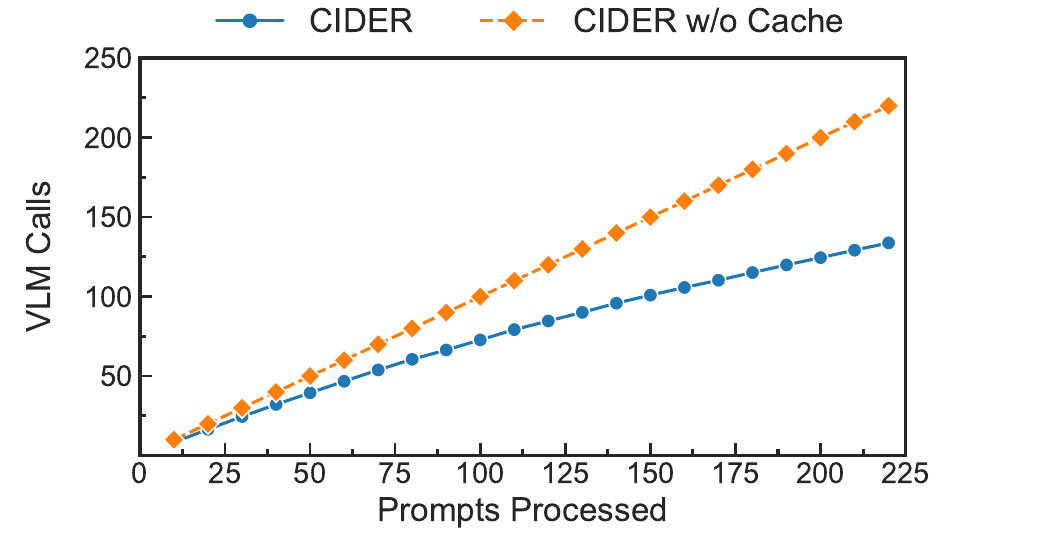}
    \caption{Ablation Study of CIDER on redirection cache.}
    \label{fig:my_diagram}
\end{figure}

\noindent
\textbf{Redirection Cache.} To evaluate the efficiency gains from the Redirection Cache, we measured the cumulative number of VLM calls required to process our dataset. To ensure robust results, we report the average over 20 independent runs with randomly shuffled data. As illustrated in Figure \ref{fig:my_diagram}, with the cache, the VLM calls volume flattens dramatically after an initial warm-up phase instead of scaling linearly. This finding demonstrates that the cache significantly reduces computational overhead, affirming its role in making CIDER both effective and economical.

\subsection{Human Evaluation.} 

We conducted a human evaluation study using graduate students from top-tier universities as expert evaluators. For a selected subset of prompts, we presented images generated by each method in a randomized and anonymous order. Evaluators were instructed to select any image(s) they deemed superior based on three criteria: (1) Bias Mitigation, (2) Image Quality, and (3) Prompt Adherence. A "win" was recorded for a method each time one of its generated images was selected. We then calculated the overall win rate for each method by aggregating these selections across all evaluators and prompts. Figure \ref{fig:human_eval} demonstrates the success of our method, providing strong substantiation for our central claims.

\section{Conclusion}
\label{sec:majhead}
In this work, we present a systematic solution to brand bias, a critical yet under-explored issue in text-to-image models. We begin by formalizing brand bias within a causal framework and introduce CIDER, a novel, model-agnostic framework that performs an inference-time causal intervention by intelligently refining prompts. Our experiments first confirm the pervasiveness of this bias across leading T2I models and then demonstrate that CIDER effectively eliminates both explicit and implicit biases without compromising image quality. Our work enhances the understanding of commercial biases in generative AI and contributes to the development of fairer, more legally compliant, and trustworthy systems.





\bibliographystyle{IEEEbib}
\bibliography{strings,refs}

\end{document}